\documentclass[journal]{IEEEtran}

\usepackage{cite}
\usepackage{array}
\usepackage{hyperref}       % hyperlinks
\usepackage{url}            % simple URL typesetting
\usepackage{booktabs}       % professional-quality tables
\usepackage{amsfonts}       % blackboard math symbols
\usepackage{nicefrac}       % compact symbols for 1/2, etc.
\usepackage{microtype}      % microtypography

\usepackage{graphicx} % more modern
\usepackage{subfigure} 
\usepackage{amsmath}

\DeclareMathOperator{\softmax}{softmax}
\DeclareMathOperator*{\argmax}{argmax} % no space, limits underneath in displays
\usepackage{tabularx}
\usepackage{multirow}
\usepackage[ruled,vlined]{algorithm2e}

\usepackage{booktabs}
\newcommand{\ra}[1]{\renewcommand{\arraystretch}{#1}}

\usepackage{lipsum}
\usepackage{dblfloatfix}

\begin{document}
	
\title{Auto-clustering Output Layer: Automatic Learning of Latent Annotations in Neural Networks}
\author{\IEEEauthorblockN{Ozsel Kilinc,~\IEEEmembership{Student Member,~IEEE,} Ismail Uysal,~\IEEEmembership{Member,~IEEE}}}
	
\markboth{Submitted to IEEE Transactions on Neural Networks and Learning Systems,~2017}%
{Shell \MakeLowercase{\textit{et al.}}: Bare Demo of IEEEtran.cls for IEEE Transactions on Magnetics Journals}	

\maketitle

\begin{abstract}
	
In this paper, we discuss a different type of semi-supervised setting: a coarse level of labeling is available for all observations but the model has to learn a fine level of latent annotation for each one of them. Problems in this setting are likely to be encountered in many domains such as text categorization, protein function prediction, image classification as well as in exploratory scientific studies such as medical and genomics research. We consider this setting as simultaneously performed supervised classification (per the available coarse labels) and unsupervised clustering (within each one of the coarse labels) and propose a novel output layer modification called auto-clustering output layer (ACOL) that allows concurrent classification and clustering based on Graph-based Activity Regularization (GAR) technique. As the proposed output layer modification duplicates the softmax nodes at the output layer for each class, GAR allows for competitive learning between these duplicates on a traditional error-correction learning framework to ultimately enable a neural network to learn the latent annotations in this partially supervised setup. We demonstrate how the coarse label supervision impacts performance and helps propagate useful clustering information between sub-classes. Comparative tests on three of the most popular image datasets MNIST, SVHN and CIFAR-100 rigorously demonstrate the effectiveness and competitiveness of the proposed approach.

\end{abstract} 
		
%\begin{IEEEkeywords}
%% IEEE, IEEEtran, IEEE Transactions on Magnetics, journal, \LaTeX, magnetics, paper, template.
%\end{IEEEkeywords}
		
\IEEEpeerreviewmaketitle
		
\section{Introduction}

\IEEEPARstart{C}{ombination} of supervised and unsupervised learning have resulted in many fruitful developments throughout the machine learning literature. Among many achievements, unsupervised feature learning where unsupervised training is used as a pre-training stage for initializing hidden layer parameters \cite{hinton2006fast}, was the first method to succeed in the training of fully connected architectures and played a key role in igniting the third wave of machine learning research by creating a paradigm shift we now call deep learning \cite{Goodfellow-et-al-2016}. 

In current literature, different kinds of approaches exist to combine supervised and unsupervised learning. In this context, \textit{semi-supervised} term is frequently used to specify certain applications where a large number of observations exist with only a small subset having ground-truth labels. Solutions suggested for these applications seek the ways of exploiting the unlabeled data to improve the model generalization. There have been significant developments recently in this field. Following the introduction of Bayesian inference to the conventional autoencoder architecture \cite{KingmaW13, RezendeMW14}, these variational autoencoders have been proven to make deep generative models highly competitive for semi-supervised learning \cite{KingmaMRW14, MaaloeSSW16}. Virtual adversarial training \cite{MiyatoMKNI15} motivated by Generative Adversarial Nets \cite{GoodfellowPMXWOCB14} and denoising autoencoder variant named Ladder networks \cite{Valpola14} have also been successfully employed for semi-supervised learning problems \cite{RasmusBHVR15}. On the other hand,  \cite{Kilinc2017Gar} have recently proposed a scalable and efficient graph-based method that is natural to the operational mechanism of deep neural networks and reported competitive performance with respect to other approaches. 

\begin{figure}[ht]
	\begin{center}
		\centerline{\includegraphics[width=\columnwidth,trim={3cm 5.5cm 5.5cm 2cm},clip]{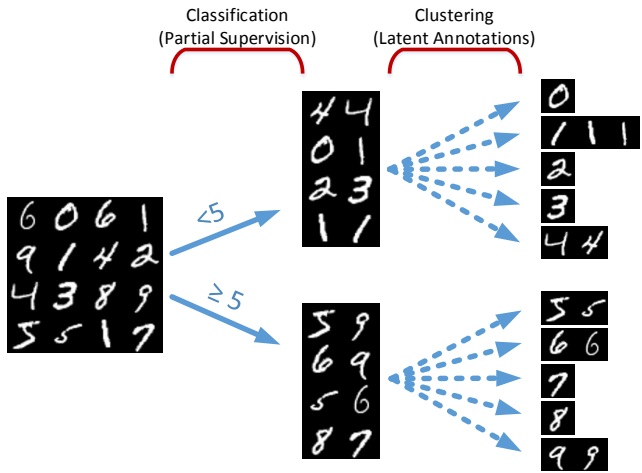}}
		\caption{In semi-supervised learning problems discussed in this paper, a coarse level of labeling is available for all observations but the model has to learn a fine level of latent annotation for each one of them. We propose a novel approach that considers these problems as simultaneously performed supervised classification (per provided parent-classes) and unsupervised clustering (within each parent) tasks and devise a framework to enable a neural network to learn the latent annotations in this partially supervised setup.}
		\label{fig:motivation1}
	\end{center}
\vskip -0.2 in
\end{figure}

In this paper, we consider a different kind of semi-supervised setting in which a coarse level of labeling is available for all observations but the model still needs to learn a fine level of latent annotation for each one of them. Provided labeling can be interpreted as parent-class information and latent annotations to be explored can be conceived as the sub-classes. Since provided partial supervision does not involve any explicit information, sub-class exploration is considered an unsupervised task. Therefore, the overall learning procedure can be considered semi-supervised. For clarification, let us assume that we are given a dataset of hand-written digits such as MNIST \cite{lecun1998mnist} where the overall task is complete categorization of each digit, but the only available supervision is whether a digit is smaller or greater than 5, as visualized in Figure \ref{fig:motivation1}. While being trained to categorize each example as a member of parent-classes, $\{0,1,2,3,4\}$ or $\{5,6,7,8,9\}$, the model also needs to learn to distinguish the digits, sub-classes under the same parent from each other. Since provided labeling involve no explicit information about the difference between the samples of digit 0 and the samples of digit 1, their separation is performed as an unsupervised task. As we use a partial supervision to help overall categorization, the entire procedure is semi-supervised.

Outside the neural network literature, the learning of latent variable models when supervision for more general classes than those of interest is provided has previously been studied. In natural language processing (NLP) field, following the introduction of Latent Dirichlet Allocation (LDA), a completely unsupervised algorithm that models
each document as a mixture of topics \cite{BleiNJ03}, several modifications have been proposed to incorporate supervision \cite{BleiM07, Lacoste-JulienSJ08, RamageHMG09, RamageHNM09}. These ideas have also been extended for simultaneous image classification and annotation \cite{WangBL09, RasiwasiaV13}.

From the viewpoint of parent/sub-class interpretation of the provided supervision and latent annotations, an analogy can be established between the semi-supervised problems discussed in this paper and the hierarchical classification tasks in the literature. Hierarchical classification has previously been studied in neural networks \cite{Xiao2007}; however, proposed approaches consider the completely supervised case where sub-class labels are known even at relatively smaller numbers than parent labels. The literature about hierarchical classification is scattered across very different application domains such as text categorization, protein function prediction, music genre classification and image classification \cite{SillaF11}. 

Given a partial supervision, latent annotation learning is also a common problem in these domains as well as in exploratory scientific research such as medical and genomics \cite{bair2013semi}, as the tasks in these domains naturally involve both already-explored (hence labeled) classes and extraction of not-yet-explored (hence hidden) patterns. In this paper, we propose a novel approach that considers these problems as simultaneously performed supervised classification (per provided parent-classes) and unsupervised clustering (within each parent) tasks and devise a framework to enable a neural network to learn the latent annotations in this partially supervised setup. This framework involves a novel output layer modification called auto-clustering output layer (ACOL) that allows concurrent classification and clustering tasks where clustering is performed according to Graph-based Activity Regularization (GAR) technique recently proposed in \cite{Kilinc2017Gar}. ACOL duplicates the softmax nodes at the output layer and GAR allows for competitive learning between these duplicates on a traditional error-correction learning framework. 

This paper is organized as follows. Next section briefly summarizes the activity regularization proposed in \cite{Kilinc2017Gar} which we adopt as the objective of the unsupervised portion of the training. In the third section, we describe the proposed output layer modification and its integration with GAR technique. Experimental results are presented in the fourth section and the paper is concluded with final remarks and possible directions for future work.
		
\section{Related Work}

GAR is a scalable and efficient graph-based approach which is originally proposed for the classical type of semi-supervised learning problems where a large number of observations with only a small subset of corresponding labels exist. In this paper, we adopt the same regularization terms and show that these terms can be employed to reveal the latent annotations in a supervised setup through ACOL. Before describing ACOL, this section briefly summarizes the activity regularization proposed in \cite{Kilinc2017Gar}. 

Unlike conventional graph-based methods estimating the adjacency matrix (which describes the similarity between the observations) using an auxiliary algorithm such as nearest neighbor or auxiliary external knowledge, GAR proposes to infer the adjacency through the actual predictions of a neural network model initialized by a supervised pretraining using the available labeled observations. After pretraining, predictions of the network, $\boldsymbol{B}$, for all $m$ examples are obtained as an $m \times n$ matrix and the adjacency of the examples are then inferred by $m \times m$ symmetric matrix $\boldsymbol{M}$ defined as
\begin{equation}
\boldsymbol{M}:=\boldsymbol{BB}^T
\end{equation}
where $n$ is the number of output classes and $B_{ij}$ is the probability of the $i$\textsuperscript{th} example belonging to $j$\textsuperscript{th} class. During the subsequent unsupervised portion of the training, label information is propagated across the graph $\mathcal{G}_\mathcal{M}$ described by $\boldsymbol{M}$. 

For a scalable and efficient optimization, rather than explicitly regularizing the matrix $\boldsymbol{M}$, GAR defines the objective of the unsupervised training through the regularization of $n \times n$ symmetric matrix $\boldsymbol{N}$ defined as
\begin{equation}
\boldsymbol{N}:=\boldsymbol{B}^T\boldsymbol{B}
\end{equation}
in order to become the identity matrix. Let $\boldsymbol{v}$ be a $1 \times n$ vector representing the diagonal entries of $\boldsymbol{N}$ such that $\boldsymbol{v}:=\begin{bmatrix}N_{11} & N_{22} & \dots  & N_{nn}\end{bmatrix}$ and $\boldsymbol{V}$ be defined as $n \times n$ symmetric matrix such that $\boldsymbol{V}:=\boldsymbol{v}^T\boldsymbol{v}$. Then, \textit{affinity} term penalizing the non-zero off-diagonal entries of $\boldsymbol{N}$ is defined as 
\begin{equation}
\label{affinity}
\textit{Affinity} = \alpha\big(\boldsymbol{B}\big) :=\frac{\sum\limits_{i \ne j}^n{N_{ij}}}{(n-1)\sum\limits_{i = j}^n{N_{ij}}}
\end{equation}
and \textit{balance} term attempting to equalize diagonal entries is written as 
\begin{equation}
\label{balance}
\textit{Balance} = \beta\big(\boldsymbol{B}\big) := \frac{\sum\limits_{i \ne j}^n{V_{ij}}}{(n-1)\sum\limits_{i = j}^n{V_{ij}}}
\end{equation}
It has been shown that as the matrix $\boldsymbol{N}$ turns into the identity matrix, $\mathcal{G}_\mathcal{M}$ becomes a disconnected graph including $n$ disjoint subgraphs each of which is $\nicefrac{m}{n}$-regular. This indicates that the strong adjacencies in the matrix $\boldsymbol{M}$ get stronger, weak ones diminish and each label is propagated to $\nicefrac{m}{n}$ examples through the strong adjacencies. Ultimately, $\boldsymbol{M}$ yields that $\boldsymbol{B}$ represents the optimal embedding.

Consider a neural network with $L-1$ hidden layers where $l$ denotes the individual index for each layer such that $l \in \{0,...,L\}$. Let $\boldsymbol{Y} ^{(l)}$ denote the output of the nodes at layer $l$. $\boldsymbol{Y} ^{(0)}=\boldsymbol{X}$ is the input and $f(\boldsymbol{X})=f^{(L)}(\boldsymbol{X})=\boldsymbol{Y}^{(L)}=\boldsymbol{Y}$ is the output of the entire network. $\boldsymbol{W} ^{(l)}$ and $\textbf{b}^{(l)}$ are the weights and biases of layer $l$, respectively. Then, the feedforward operation of the neural networks can be written as 
\begin{equation}
\label{neuralnetwork}
f^{(l)}\big(\boldsymbol{X}\big) = 
\boldsymbol{Y}^{(l)} = 
h^{(l)}\big(\boldsymbol{Y}^{(l-1)}\boldsymbol{W}^{(l)} + \boldsymbol{b}^{(l)}\big)
\end{equation}
where $h^{(l)}$(.) is the activation function applied at layer $l$. Using this notation, let us define the activities at the input of softmax layer as $\boldsymbol{Z}$ such that  
\begin{equation}
\label{GARZ}
\boldsymbol{Z} := 
\boldsymbol{Y}^{(L-1)}\boldsymbol{W}^{(L)} + \boldsymbol{b}^{(L)}
\end{equation}
Rather than using the probabilistic output of the softmax nodes i.e. $\boldsymbol{Y} = \softmax(\boldsymbol{Z})$, GAR technique applies the regularization over the positive part of the activities at their inputs such that 
\begin{equation}
\label{garpredictions}
g\big(\boldsymbol{X}\big) = 
\boldsymbol{B} := 
\max{\big(\boldsymbol{0}, \boldsymbol{Z}\big)}
\end{equation}
for an easier optimization task. 
Then, the overall unsupervised regularization loss proposed by GAR ultimately becomes
\begin{equation}
\label{unsupervised_objective}
\mathcal{U}\big(\boldsymbol{B}\big)= c_{\alpha}\alpha\big(\boldsymbol{B}\big) + c_{\beta}\big(1-\beta\big(\boldsymbol{B}\big)\big) + c_F||\boldsymbol{B}||^2_F
\end{equation}
where $||\boldsymbol{B}||_F$ corresponds to the Frobenius norm for $\boldsymbol{B}$ employed to limit the denominators of both \textit{affinity} and \textit{balance} terms not to diminish their effects and $c_{\alpha}, c_{\beta}, c_F$ are the weighting coefficients. 

\section{Auto-clustering Output Layer}
\subsection{Output Layer Modification}

Neural networks define a family of functions parameterized by weights and biases which define the relation between inputs and outputs. In multi-class categorization tasks, outputs correspond to class labels, hence in a typical output layer structure there exists an individual output node for each class. An activation function, such as softmax is then used to calculate normalized exponentials to convert the previous hidden layer's activities, i.e. scores, into probabilities.

\begin{figure}[ht]
	\begin{center}
		\centerline{\includegraphics[width=\columnwidth,trim={1cm 1.1cm 1.8cm 1.2cm},clip]{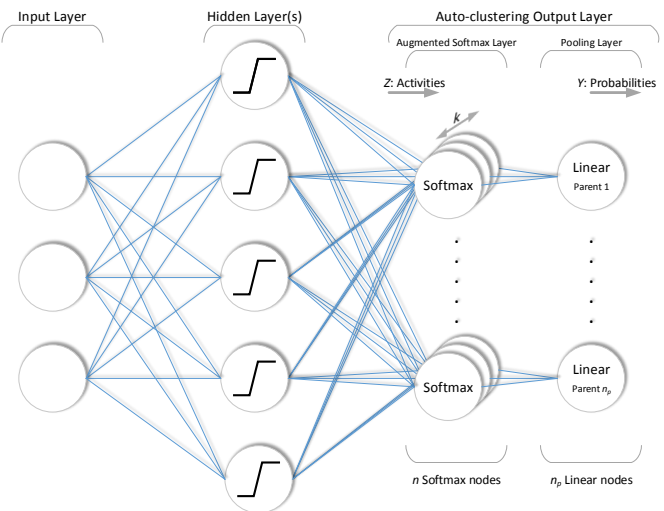}}
		\caption{Neural network structure with the ACOL. Each softmax node corresponds to an individual sub-class of a parent, i.e. annotation. During feedforward operation of the network, pooling layer calculates final parent-class predictions through sub-class probabilities.}
		\label{fig:clustering}
	\end{center}
	\vskip -0.2 in
\end{figure}

Unlike traditional output layer structure, ACOL defines more than one softmax node ($k$ duplicates) per parent-class. Outputs of $k$ duplicated softmax nodes that belong to the same parent are then combined in a subsequent pooling layer for the final prediction. Training is performed in the configuration shown in Figure~\ref{fig:clustering} where $n_p$ is the number of parent-classes. This might look like a classifier with redundant softmax nodes. However, duplicated softmax nodes of each parent are specialized using GAR throughout the training in a way that each one of $n=n_pk$ softmax nodes represent an individual sub-class of a parent, i.e. annotation.

ACOL does not change feedforward and backpropagation mechanisms of the network drastically. During feedforward operation of the network, pooling layer calculates final parent-class predictions through sub-class probabilities. Pooling does not affect backpropagation in terms of derivatives and ACOL behaves in a similar fashion to traditional output layer. However, labels are now implicitly applied to multiple softmax nodes each representing an individual sub-class under the same parent. In other words, even if the labels are provided as one-hot encoded vector at the output, due to the pooling layer, it turns into $k$-hot encoded vector at the augmented softmax layer. $k$ softmax nodes simultaneously receive the error between the label and the prediction and then backpropagate it towards the previous hidden layers.

This structure carries the ability to learn latent annotations as ACOL introduces extra trainable weights between the previous hidden layer and itself. Each softmax node is connected to the previous hidden layer through non-shared weights. Due to random initialization, these weights may ultimately converge to different values at the end of training and duplicated softmax nodes may be specialized for only a subset of the samples of that parent-class. However, this mechanism is completely uncontrollable. Furthermore, during the weight updates, if any one of the duplicated softmax nodes get activated to generate a significantly lower error, through the pooling layer, this will also eliminate the backpropagated error to other $k-1$ softmax nodes of that parent. Therefore, not only the one reducing the error, but all $k$ duplicated softmax nodes diminish backpropagating the error to previous layers. That is to say, without any additional mechanism, there is no actual competition between the duplicated softmax nodes of a parent.

Therefore, we adopt GAR objective defined in (\ref{unsupervised_objective}) as the unsupervised regularization term to create competition between the duplicates which ultimately results in specialized but equally-active softmax nodes each representing a latent annotation within a parent. Following subsection mathematically describes ACOL and its collaboration with GAR.  

\subsection{Mathematical Description and GAR Integration}

In a neural network with ACOL, due to the subsequent pooling layer, (\ref{GARZ}) is modified as 
\begin{equation}
\boldsymbol{Z} := \boldsymbol{Y}^{(L-2)}\boldsymbol{W}^{(L-1)} + \boldsymbol{b}^{(L-1)}
\end{equation}
and now $\boldsymbol{Z}$ corresponds to $m \times n$ matrix representing the activities going into the augmented softmax layer. Recall that $n$ is the total number of all softmax nodes at the augmented softmax layer such that $n=n_pk$, where $n_p$ is the number of parent-classes and $k$ is the clustering coefficient of ACOL.
Then, the output of the ACOL applied network can be written in terms of $\boldsymbol{Z}$ as 
\begin{equation}
\label{acoloutput}
f\big(\boldsymbol{X}\big) = 
\boldsymbol{Y} =  
h^{(L)}\bigg(h^{(L-1)}\big(\boldsymbol{Z}\big)\boldsymbol{W}^{(L)} + \boldsymbol{b}^{(L)}\bigg)
\end{equation}
where $\boldsymbol{Y}$ is an $m \times n_p$ matrix whose cell $Y_{ij}$ represents the probability of $i$\textsuperscript{th} example belonging to $j$\textsuperscript{th} parent. Since $h^{(L-1)}(.)$ and $h^{(L)}(.)$ respectively correspond to softmax and linear activation functions and $\boldsymbol{b}^{(L)} := \boldsymbol{0}$ for ACOL networks, then (\ref{acoloutput}) further simplifies into 
\begin{equation}
%f\big(\boldsymbol{X}\big) = 
\boldsymbol{Y} =  
\softmax\big(\boldsymbol{Z}\big)\boldsymbol{W}^{(L)}
\end{equation}
where $\boldsymbol{W}^{(L)}$ (hereafter will be denoted as $\boldsymbol{W}$ for simplicity) is an $n \times n_p$ matrix representing the constant weights between the augmented softmax layer and the pooling layer such that 
\begin{equation}
\boldsymbol{W}:=
\boldsymbol{W}^{(L)}=
\begin{bmatrix}
\boldsymbol{I}_{n_p} \\
\boldsymbol{I}_{n_p} \\
\vdots \\
\boldsymbol{I}_{n_p}
\end{bmatrix}
\end{equation}
and simply sums up the output probabilities of the softmax nodes belonging to the same parent. Since the output of the augmented softmax layer is already normalized, no additional averaging is needed at the pooling layer and summation alone is sufficient to calculate final parent-class probabilities. 

Recalling that GAR is applied to the positive part of activities going into the augmented softmax layer, i.e. $\boldsymbol{B} := \max{(\boldsymbol{0}, \boldsymbol{Z})}$, the overall objective cost function of the training can be written as
\begin{equation}
\label{overallobj}
%\mathcal{L}\big(f\big(\boldsymbol{X}\big), \boldsymbol{t}\big) + 
%\mathcal{U}\big(g\big(\boldsymbol{X}\big)\big) = 
%\mathcal{L}\big(\boldsymbol{Y}, \boldsymbol{t}\big) +
%\mathcal{U}\big(\boldsymbol{B}\big)
\mathcal{L}\big(\boldsymbol{Y}, \boldsymbol{t}\big) + c_{\alpha}\alpha\big(\boldsymbol{B}\big) + c_{\beta}\big(1-\beta\big(\boldsymbol{B}\big)\big) + c_F||\boldsymbol{B}||^2_F
\end{equation}
where $\mathcal{L}(.)$ is the supervised log loss function and $\boldsymbol{t}=[t_1,...,t_{m}]^T$ is the vector of provided parent-class labels such that $t_i \in \{1,...,n_p\}$. Also, recall that $\alpha(.)$ and $\beta(.)$ are the unsupervised regularization terms respectively defined in (\ref{affinity})-(\ref{balance}) and $||\boldsymbol{B}||_F$ corresponds to the Frobenius norm for $\boldsymbol{B}$.

\subsection{Training and Annotation Assignment}

Training of the proposed framework is performed according to simultaneous supervised and unsupervised updates resulting from the objective function given in (\ref{overallobj}). We adopt stochastic gradient descent in mini-batch mode \cite{bottou10sgd} for optimization. Algorithm 1 below describes the entire training procedure. 
\begin{algorithm}[t]
	\begin{small}
		\DontPrintSemicolon
		\SetKwFunction{proc}{proc}
		\SetKwInOut{Input}{Input}
		
		\Input{$\boldsymbol{X}=[\boldsymbol{x}_1,...,\boldsymbol{x}_{m}]^T$, \\
			$\boldsymbol{t}=[t_1,...,t_{m}]^T$ such that $t_i \in \{1,...,n_p\}$, \\
			batch size $b$, weighing coefficients $c_{\alpha}, c_{\beta}, c_F$}
		\Repeat{\textnormal{stopping criteria is met}}
		{
			$ \big\{ (\boldsymbol{\acute{X}}_1, \boldsymbol{\acute{t}}_1),..., (\boldsymbol{\acute{X}}_{\nicefrac{m}{b}}, \boldsymbol{\acute{t}}_{\nicefrac{m}{b}})\big\} \longleftarrow (\boldsymbol{X},\boldsymbol{t})$ \\
			\tcp{Shuffle and create batch pairs}
			\For{$i \gets 1$ \textbf{to} $\nicefrac{m}{b}$}{
				Take $i$\textsuperscript{th} pair $(\boldsymbol{\acute{X}}_i, \boldsymbol{\acute{t}}_i)$ \\
				Forward propagate for $\boldsymbol{\acute{Y}}_i=f(\boldsymbol{\acute{X}}_i)$ and  
				$\boldsymbol{\acute{B}}_i=g(\boldsymbol{\acute{X}}_i)$ \\
				Take a gradient step for $\mathcal{L}\big(\boldsymbol{\acute{Y}}_i,\boldsymbol{\acute{t}}_i\big) + c_{\alpha}\alpha\big(\boldsymbol{\acute{B}}\big) + c_{\beta}\big(1-\beta\big(\boldsymbol{\acute{B}}\big)\big) + c_F||\boldsymbol{\acute{B}}||^2_F$
			}
		}
		
		\caption{Model training}
		
	\end{small}
\end{algorithm}

\begin{figure}[h]
	\begin{center}
		\centerline{\includegraphics[width=\columnwidth,trim={1cm 1.1cm 1.8cm 1.2cm},clip]{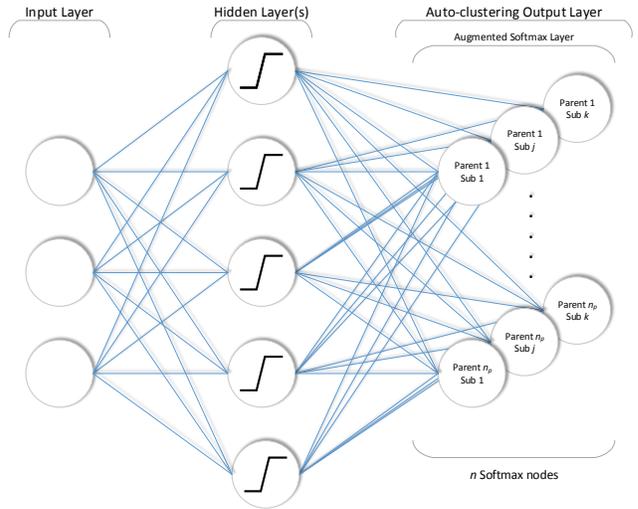}}
		\caption{After training, the pooling layer is simply disconnected. The rest of the network with trained weights is used to obtain the assigned annotations. This operation can be described as $y_i :=  \argmax_{1\le j \le n}Z_{ij}$ where $y_{i_p} :=  (y_i-1) \bmod{n_p} + 1$ and $y_{i_s} :=  (y_i-1) \backslash n_p + 1$ are corresponding parent and sub-class indices, respectively.}
		\label{fig:clustering_red}
	\end{center}
\vskip -0.2 in
\end{figure}

After training phase is completed, network is simply truncated by completely disconnecting the pooling layer as shown in Figure~\ref{fig:clustering_red} and the rest of the network with trained weights is used to assign the annotations to each example. This assignment can be described as
\begin{equation}
\label{assignedannotation}
y_i :=  \argmax_{1\le j \le n}Z_{ij}
\end{equation}
where $y_i$ is the annotation assigned to $i$\textsuperscript{th} example such that $y_{i_p} :=  (y_i-1) \bmod{n_p} + 1$ and $y_{i_s} :=  (y_i-1) \backslash n_p + 1$ are corresponding parent (learned through the provided supervision) and sub-class (learned through the unsupervised exploration) indices, respectively. 

\subsection{Graph Interpretation of the Proposed Framework}

\begin{figure*}[b]
	\begin{center}
		\centerline{\includegraphics[width=\linewidth,trim={0.5cm 1.1cm 1cm 1cm},clip]{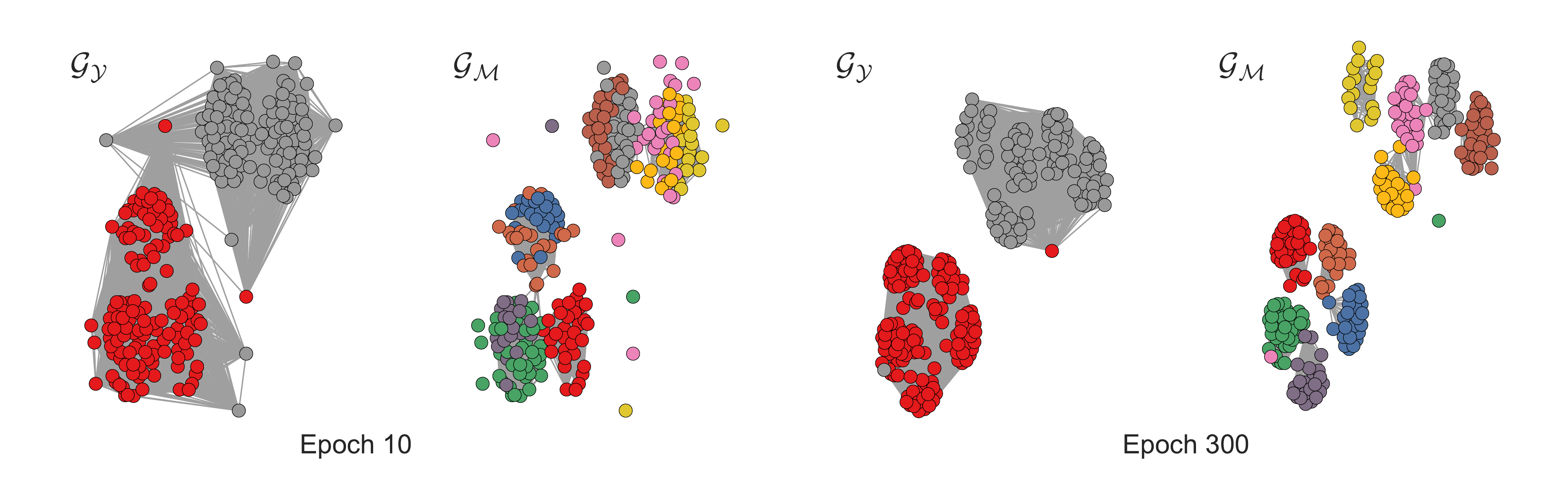}}
		\caption{Visualizations of the graph $\mathcal{G}_\mathcal{Y}$ and its spanning subgraph $\mathcal{G}_\mathcal{M}$ for randomly chosen 250 test examples from MNIST. Colored circles denote the ground-truths for the vertices, i.e. examples, and gray lines denote the edges, i.e. weighted connections between the examples representing their similarity. Note that, for vertices in graph $\mathcal{G}_\mathcal{Y}$, there are two different colors indicating true parent-class label assigned in (\ref{mnist2class}), albeit ten different colors indicating the real $digit$ identity for vertices in graph $\mathcal{G}_\mathcal{M}$. As training continues, provided supervision turns graph $\mathcal{G}_\mathcal{Y}$ into a disconnected graph of $n_p=2$ disjoint subgraphs and implicitly propagates to graph $\mathcal{G}_\mathcal{M}$ in a way that $\mathcal{G}_\mathcal{M}$ becomes a disconnected graph of $n=n_pk=10$ disjoint subgraphs where $k=5$ for this experiment. This figure is best viewed in color.}
		\label{fig:mnist_graph}
	\end{center}
\end{figure*}

GAR regularizers, \textit{affinity} and \textit{balance}, have originally been proposed for the classical type of semi-supervised learning problems where the number of labeled observations is much smaller than the number of unlabeled observations, but all existing classes are equally represented by the available labels even at limited numbers. These two terms are used to propagate these labels to unlabeled observations across the graph $\mathcal{G}_\mathcal{M}$ which is defined by $\boldsymbol{B}\boldsymbol{B}^T$. Unlike these problems, in this paper, we consider a different case in which a coarse level of labeling is available for all observations but the model has to explore a fine level of latent annotations using this partial supervision. Therefore, graph interpretation of the proposed framework in this paper is rather different than the one described in \cite{Kilinc2017Gar}.

To better understand how the provided partial supervision is propagated in order to reveal latent annotations, let us first note that even though $\softmax(\boldsymbol{Z})$ produces probabilistic output, it is an increasing function of $\boldsymbol{Z}$ similar to $\max{\big(\boldsymbol{0}, \boldsymbol{Z}\big)}$. Assuming $\softmax(\boldsymbol{Z}) \approx  \max{\big(\boldsymbol{0}, \boldsymbol{Z}\big)}$ allows us to explicitly express $\boldsymbol{Y}$ in terms of $\boldsymbol{B}$ such that

\begin{equation}
\boldsymbol{Y} \approx  \boldsymbol{B}\boldsymbol{W}
\end{equation}
Noting that $\boldsymbol{W}\boldsymbol{W}^T$ is an $n \times n$ symmetric matrix such that
\begin{equation}
\boldsymbol{W}\boldsymbol{W}^T = 
\begin{bmatrix}
\boldsymbol{I}_{n_p} & \boldsymbol{I}_{n_p} & \dots & \boldsymbol{I}_{n_p}\\
\boldsymbol{I}_{n_p} & \boldsymbol{I}_{n_p} & \dots & \boldsymbol{I}_{n_p}\\
\vdots & \vdots & \ddots & \vdots \\
\boldsymbol{I}_{n_p} & \boldsymbol{I}_{n_p} & \dots & \boldsymbol{I}_{n_p}
\end{bmatrix}
\end{equation}
this assumption helps us visualize graph $\mathcal{G}_\mathcal{M}$ (whose edges are described by $\boldsymbol{B}\boldsymbol{B}^T$) as the spanning subgraph of $\mathcal{G}_\mathcal{Y}$ (whose edges described by $\boldsymbol{Y}\boldsymbol{Y}^T = \boldsymbol{BW}\boldsymbol{W}^T\boldsymbol{B}^T$). In other words, these two graphs are made up of the same vertices. However, while propagating the supervised adjacency introduced by $\mathcal{G}_\mathcal{Y}$ across $\mathcal{G}_\mathcal{M}$, GAR regularizers eliminate some of the edges of $\mathcal{G}_\mathcal{Y}$ from $\mathcal{G}_\mathcal{M}$ in a way that $\mathcal{G}_\mathcal{M}$ ultimately becomes a disconnected graph of $n$ disjoint subgraphs. This propagation/elimination process is better explained in the following section through empirical demonstration on real data along with the impact of provided supervision on the exploration of latent annotations.

\section{Experimental Results}

The models have been implemented in Python using Keras \cite{chollet2015keras} and Theano \cite{Theano}. Open source code is available at \hyperref{http://github.com/ozcell/lalnets}{}{}{http://github.com/ozcell/lalnets} that can be used to reproduce the experimental results obtained on the three image datasets, MNIST \cite{lecun1998mnist}, SVHN \cite{svhn} and CIFAR-100 \cite{cifar} most commonly used by previous researchers publishing in the field of semi-supervised learning at NIPS, TNNLS and other similar venues.

All experiments have been performed on a 6-layer convolutional neural network (CNN) model. For MNIST and SVHN experiments, coefficients of GAR terms have been chosen as $c_\alpha=0.1, c_\beta=0.1$, $c_F=0.0003$ and supervision is introduced as a two parent-class classification problem, i.e. $n_p=2$, as further explained in the following sections. CIFAR-100 naturally involves two levels of labeling where $n_p=20$. For CIFAR-100, we use the same values for $c_\alpha$ and $c_\beta$ but change $c_F$ to $10^{-7}$ due to the difference in $n_p$ setting. For all experiments, we used a batch size of 128. Each experiment has been repeated for 10 times. A validation set of 1000 examples has been chosen randomly among the training set examples to determine the epoch to report the test performance, which is obtained through the examples not introduced to the model during training, as is standard. For the sake of fairness, to obtain the performances of the models used for comparison, all training examples of the datasets are used for the pretraining of autoencoder-based models, and datasets are later manually pre-divided into two subsets according to the provided supervision where two individual clusterings are performed within these subsets. The overall performances are obtained by combining the results of these two clusterings. Following \cite{jiangZTTZ16}, we evaluate test performances with unsupervised clustering accuracy given as
\begin{equation}
\label{ACC}
ACC = \max_{\mathfrak{f}\in \mathfrak{F}}\frac{\sum_{i = 1}^{m}{1\{t^*_i = \mathfrak{f}(y_i)\}}}{m}
\end{equation}
where $t^*_i$ is the ground-truth label, $y_i$ is the annotation assigned in (\ref{assignedannotation}), and $\mathfrak{F}$ is the set of all possible one-to-one mappings between assignments and labels.

\subsection{MNIST}
To empirically demonstrate the label propagation process and compare the proposed approach with other methods, we create a semi-supervised problem on MNIST by providing the parent-class supervision of whether a digit is smaller or larger than 5 such that
\begin{equation}
\label{mnist2class}
t_i =
\begin{cases}
0 & \text{if $digit < 5$} \\ 
1 & \text{otherwise}
\end{cases}
\end{equation}

\begin{figure*}[t]
	\begin{center}
		\centerline{\includegraphics[width=160mm,trim={0cm 0.3cm 0cm 0.5cm},clip]{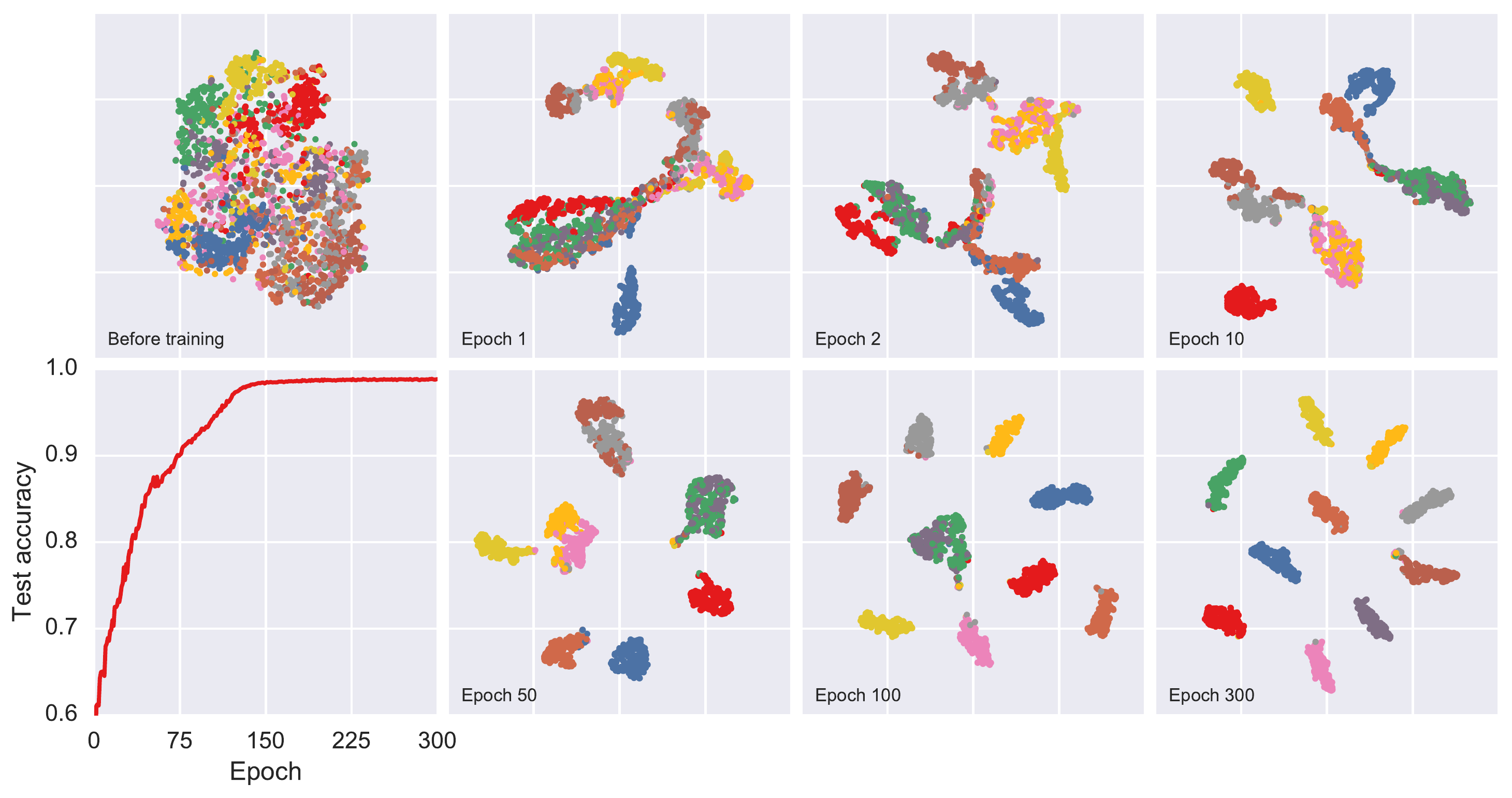}}
		\caption{t-SNE visualization of the latent space inferred by $\boldsymbol{Z}$ for randomly chosen 2000 test examples from MNIST. Color codes denote the ground-truths for the examples. Note the separation of clusters from epoch 1 to epoch 300 of the training. For reference, accuracy for the entire test set is also plotted with respect to the training epochs. This figure is best viewed in color.}
		\label{fig:mnist_tsne}
	\end{center}
	\vskip -0.2 in
\end{figure*}

Figure \ref{fig:mnist_graph} visualizes the realization of label propagation using the real predictions obtained for this problem. Colored circles denote the ground-truths for the vertices, i.e. examples, and gray lines denote the edges, i.e. weighted connections between the examples representing their similarity. Note that, for vertices in graph $\mathcal{G}_\mathcal{Y}$, there are two different colors indicating true parent-class label assigned in (\ref{mnist2class}), albeit ten different colors indicating the real $digit$ identity for vertices in graph $\mathcal{G}_\mathcal{M}$. Graph $\mathcal{G}_\mathcal{M}=(\mathcal{M},\mathcal{E})$ shares the same vertices $\mathcal{M}$ with graph $\mathcal{G}_\mathcal{Y}=(\mathcal{M},\mathcal{E}_\mathcal{Y})$, which is constructed per the provided supervision. However, $\mathcal{E}$ is a subset of $\mathcal{E}_\mathcal{Y}$ as some of the edges in graph $\mathcal{G}_\mathcal{Y}$, such as those between the examples of digit 0 and 1, are eliminated in graph $\mathcal{G}_\mathcal{M}$ due to GAR regularization terms. As training continues, provided supervision turns graph $\mathcal{G}_\mathcal{Y}$ into a disconnected graph of $n_p=2$ disjoint subgraphs and implicitly propagates to its spanning subgraph $\mathcal{G}_\mathcal{M}$ in a way that $\mathcal{G}_\mathcal{M}$ becomes a disconnected graph of $n=n_pk=10$ disjoint subgraphs where $k=5$ for this experiment.  

\begin{figure}[h!]
	\begin{center}
		\centerline{\includegraphics[width=70mm,trim={0cm 0.3cm 0cm 0.5cm},clip]{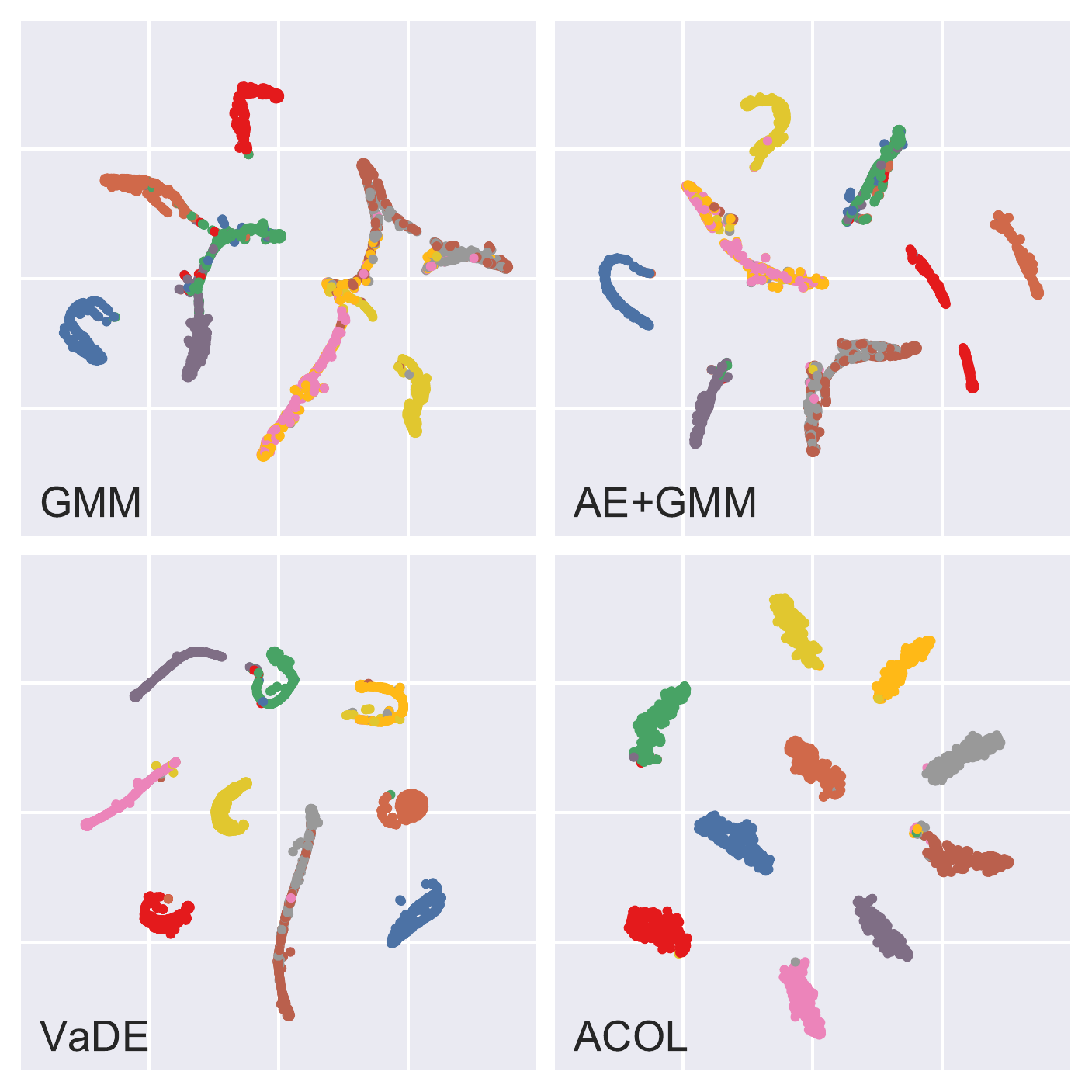}}
		\caption{t-SNE visualization of the latent spaces obtained using four different approaches for randomly chosen 2000 test examples from MNIST. In order to introduce the same two-parent supervision to other approaches, the dataset is first divided into two subsets according to the provided supervision and then distinct latent spaces obtained for each one of the subsets are combined for the final result. Color codes denote the ground-truths for the examples. Note the more definitive separation of clusters when using ACOL. This figure is best viewed in color.}
		\label{fig:mnist_tsne_comparison}
	\end{center}
	\vskip -0.2 in
\end{figure}

Figure \ref{fig:mnist_tsne} presents the t-SNE \cite{maaten2008tsne} visualization of the latent space inferred by $\boldsymbol{Z}$ for randomly chosen 2000 test examples from MNIST. From epoch 1 to epoch 300 of the training, clusters become well-separated and simultaneously the test accuracy increases. As clearly observed from this figure, using the provided partial supervision, the neural network also reveals some hidden patterns useful to distinguish the examples of different digits under the same parent-class and ultimately learns to categorize each one of the ten digits. Also for comparison, Figure \ref{fig:mnist_tsne_comparison} provides latent space visualizations obtained using three other approaches along with ACOL. In order to introduce the same two-parent supervision to other approaches, the dataset is first divided into two subsets according to the provided supervision and then distinct latent spaces obtained for each one of the subsets are combined for the final result. Table \ref{tab:mnist} summarizes the test error rates calculated using the unsupervised clustering accuracy metric given in (\ref{ACC}) for MNIST with $k=5$ setting. Results of a broad range of recent existing solutions are also presented for comparison. VaDE \cite{jiangZTTZ16}, unsupervised generative clustering framework combining Variational Autoencoders (VAE) and Gaussian Mixture Model (GMM) together, produces more competitive results with respect to other approaches as it adopts variational inference during the reconstruction process and enables the simultaneous updating of the GMM parameters and the network parameters. On the other hand, unlike VaDE and other similar approaches based on the reconstruction of the input, ACOL motivates neural networks to learn the latent space representation through the provided partial supervision, which is typically more general than the overall categorization interest. This motivation yields a better separation of the clusters in the latent space; however, its quality depends on the provided supervision as further explored in the following section. 

\begin{figure*}[b]
	\begin{center}
		\centerline{\includegraphics[width=\linewidth,trim={0cm 0cm 0cm 0.5cm},clip]{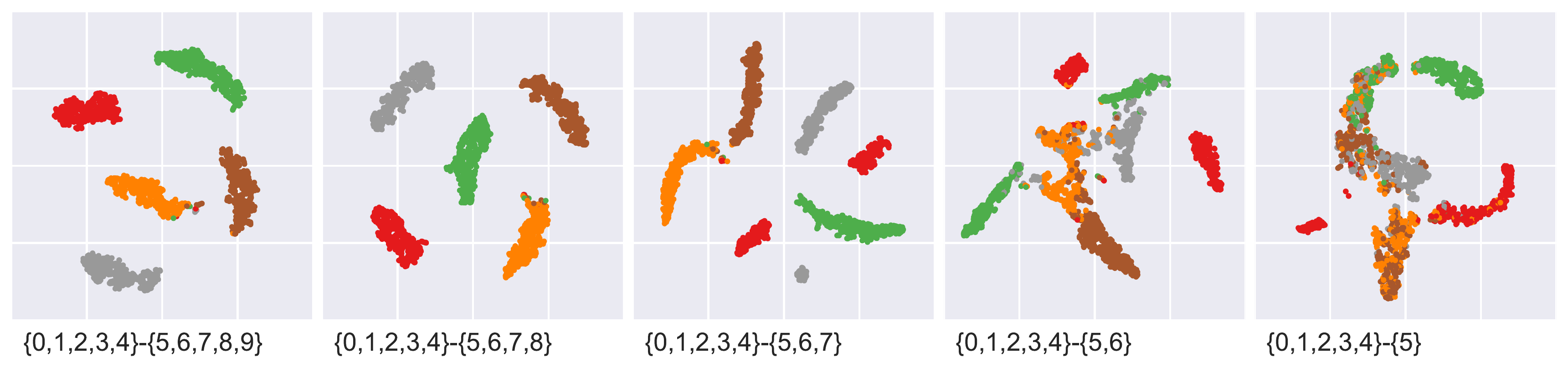}}
		\caption{t-SNE visualization of the latent spaces obtained for the first parent-class, i.e. $\{0,1,2,3,4\}$, showing the inter-parent effect of the provided supervision. In this scenario, the first parent-class is left unchanged throughout the experiment but all examples of a digit are discarded from the second parent-class in each new iteration. When the classification objective becomes more challenging due to more distinct digits in the second parent-class, the network is more capable of revealing the hidden patterns needed to better differentiate the examples of the first parent. Color codes denote the ground-truths for the examples. This figure is best viewed in color.}
		\label{fig:mnist_tsne_unbalanced}
	\end{center}
	\vskip -0.2 in
\end{figure*}

\begin{table}[h]
	\ra{1.2}
	\caption{Benchmark results for the two-parent case, i.e. whether a digit is smaller or larger than 5, on MNIST. The last row demonstrates the benchmark scores of the proposed framework in this article. Note that given values represent the test error.}
	\centering
	%\resizebox{50mm}{!} {
		\begin{tabular}{@{}lr@{}}\toprule
									& MNIST $k=5$ 				\\
			\midrule
			$k$-means				& $24.99\%\hskip 3.35em$	\\
			GMM						& $22.70\%\hskip 3.35em$	\\
			AE+$k$-means			& $21.80\%\hskip 3.35em$ 	\\
			AE+GMM					& $23.80\%\hskip 3.35em$	\\	
			VaDE \cite{jiangZTTZ16}	& $8.18\%\hskip 3.35em$ 	\\
			\midrule
			ACOL 					& $1.39\%(\pm 0.12)$		\\
			\bottomrule
			\label{tab:mnist}
		\end{tabular}
	%}
	\vskip -0.2 in
\end{table}

\subsection{MNIST - Impact of the Provided Supervision on Performance}

We perform two experiments on MNIST in order to observe the impact of provided supervision and whether useful information is propagated between parent-classes for better sub-classification. In the first experiment, we use the same supervision described in (\ref{mnist2class}) by leaving the first parent-class unchanged throughout the experiment but discarding all examples of a digit from the second parent-class in each new iteration. For all five iterations of this experiment, Figure \ref{fig:mnist_tsne_unbalanced} presents the t-SNE visualization of the latent space representation observed for randomly chosen 2000 test examples only from the first parent-class, i.e. \{0,1,2,3,4\}, and Table \ref{tab:mnist_interparent} summarizes the overall test performance across all examples included in the training. One might expect to observe better performance when the clustering problem under one of the parent-classes is simplified. On the contrary, a more challenging objective forces the network to reveal more latent patterns needed to better differentiate each of the digits. More specifically, when the second parent-class consists only of the examples of the digit 5, the network learns only to distinguish its examples from those of the first five digits. Adding the examples of digit 6, the network now has to extract more hidden patterns identifying the unique differences between the set of the digits $5, 6$ and the set of digits $0, 1, 2, 3, 4$. These extra hidden patterns also contribute to the differentiation of the digits $0, 1, 2, 3, 4$ from each other, which we identify as the \textit{inter-parent} effect of the provided supervision. 

\begin{table}[h]\centering
	\ra{1.2}
	\caption{Benchmark results for the two-parent case on MNIST to observe the inter-parent effect of the provided supervision.}
	%\resizebox{60mm}{!} {
	\begin{tabular}{@{}llr@{}}\toprule
		1\textsuperscript{st} Parent  & 2\textsuperscript{nd} Parent   & Test Error\\
		\midrule
		\{0,1,2,3,4\} & \{5,6,7,8,9\}  	& $1.39\%(\pm 0.12)$	\\
		\{0,1,2,3,4\} & \{5,6,7,8\}  	& $3.83\%(\pm 2.09)$    \\
		\{0,1,2,3,4\} & \{5,6,7\}  		& $4.04\%(\pm 0.92)$	\\
		\{0,1,2,3,4\} & \{5,6\}  		& $16.44\%(\pm 3.33)$	\\
		\{0,1,2,3,4\} & \{5\} 			& $26.55\%(\pm 2.64)$	\\
		\bottomrule
		\label{tab:mnist_interparent}
	\end{tabular}
	%	}
	\vskip -0.2 in
\end{table}

\begin{figure}[h]
	\begin{center}
		\centerline{\includegraphics[width=\columnwidth,trim={0cm 0cm 0cm 0cm},clip]{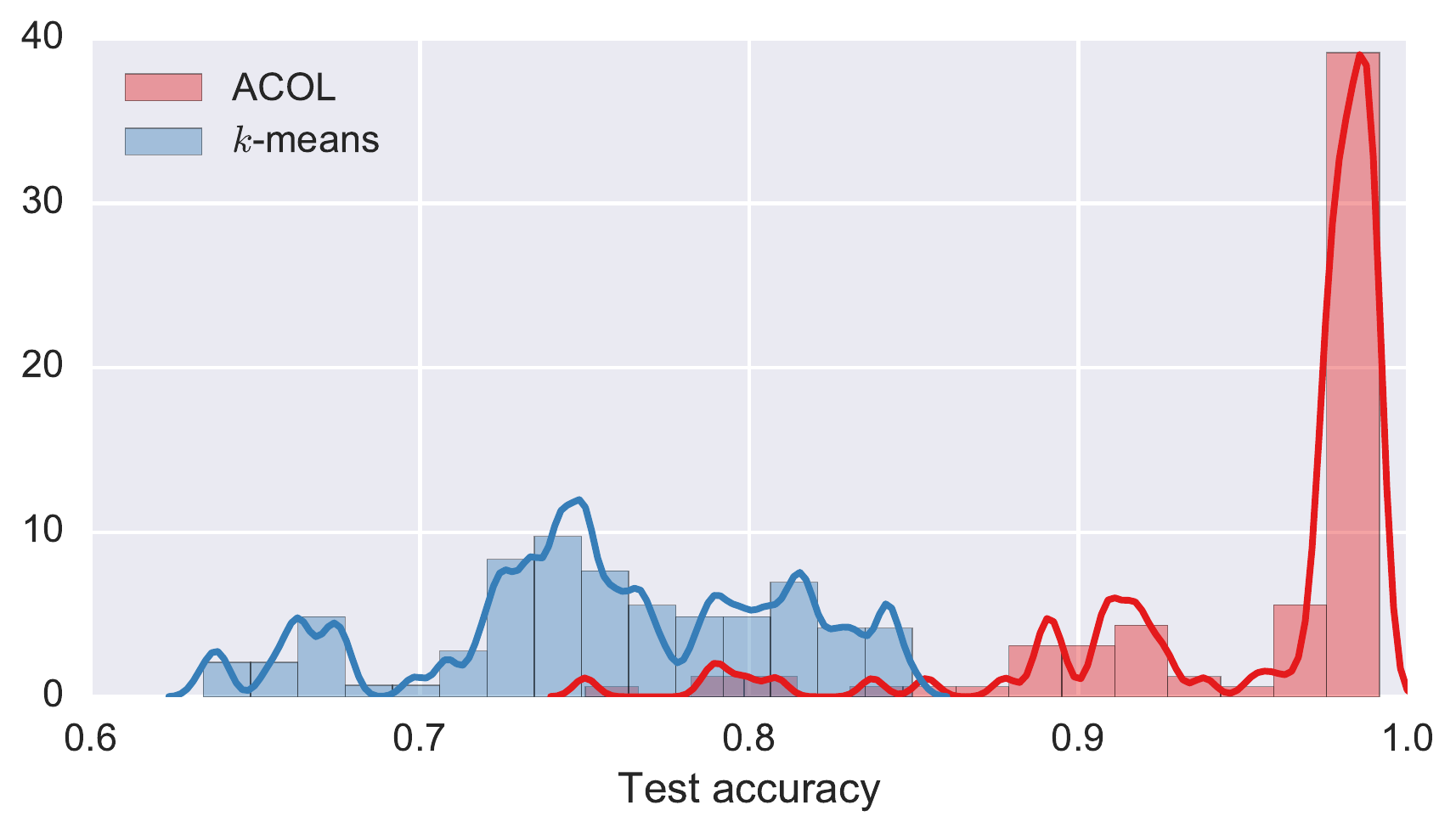}}
		\caption{Normalized histogram of the test accuracies obtained using ACOL for randomly chosen 100 two-parent supervision scenarios on MNIST. For comparison, same scenarios are also tested using $k$-means algorithm.}
		\label{fig:mnist_random}
	\end{center}
	\vskip -0.2 in
\end{figure}

In the second experiment, rather than using the rule given in (\ref{mnist2class}), we randomly assign the parent-classes. That is, examples of randomly chosen five digits are used to construct the first parent-class and those of the remaining five digits form the second one. The experiment is repeated for 100 times with a new selection of parent-classes. Histogram of the test accuracies observed for these 100 repetitions is given in Figure \ref{fig:mnist_random} and Table \ref{tab:mnist_random} summarizes the best, the median and the worst cases along with the overall average. For reference, $k$-means results obtained for the same 100 scenarios are also provided. One can observe that ACOL is less sensitive to variations in the provided supervision as majority of the iterations are concluded with a test accuracy within 1.2\% range. For $k$-means results, provided supervision only determines the difficulty of the subsequent clustering tasks, which are performed individually. However, in ACOL, there is a much more complex relation between the provided supervision and the observed performance. Even though the assigned parent-classes yield a more difficult unsupervised clustering task, ACOL can compensate this effect with the help of latent patterns learned through inter-parent comparisons due to the classification objective. Therefore, when a coarse level of labeling is available for a dataset, simultaneous classification and clustering through ACOL produces better separated and more accurate latent embeddings than individual clustering tasks within each of the known labels as ACOL enables neural networks to exploit the provided supervision. 

\begin{table}[h]\centering
	\ra{1.2}
	\caption{Test errors for worst, median and best cases among 100 two-parent supervision scenarios on MNIST.}
	%\resizebox{\columnwidth}{!} {
		\begin{tabular}{@{}lrrrr@{}}\toprule
			&Worst	& Median & Best & Mean\\
			\midrule
			$k$-means	& $36.58\%$	& $24.87\%$	& $15.04\%$ 	& $24.29\%(\pm 1.05)$\\
			ACOL		& $24.98\%$	& $2.04\%$	& $0.85\%$	& $4.49\%(\pm 1.04)$ \\
			\bottomrule
			\label{tab:mnist_random}
		\end{tabular}
	%}
	\vskip -0.2 in
\end{table}

\subsection{SVHN and CIFAR-100}

We also test the proposed approach for more realistic and challenging scenarios using SVHN and CIFAR-100 datasets. For SVHN, we adopt the same supervision defined in (\ref{mnist2class}). On the other hand, CIFAR-100 dataset naturally defines two levels of labeling. Each image has a coarse label indicating the \textit{superclass} to which it belongs, such as \textit{trees} and also a fine label indicating the \textit{class} to which it belongs, such as \textit{maple}. The 100 \textit{classes} in the CIFAR-100 are grouped into 20 \textit{superclasses}. For the CIFAR-100 experiment, coarse labels are provided as the supervision and fine labels are targeted as the annotations to be observed. Table \ref{tab:svhn} and \ref{tab:cifar} respectively summarize the test performances obtained for SVHN and CIFAR-100 datasets with two different $k$ settings. Results of a broad range of other approaches in the current literature are also presented for comparison.

It is worth noting that, unlike MNIST and SVHN, the proposed approach suffers from a limitation in CIFAR-100 dataset. That is, the parent-wise classification performance of the model affects the accuracy of assigned annotations. In MNIST and SVHN datasets, once trained with the provided parent-classes, the networks generalize well to the test sets based on this supervision. Hence, when the sub-class annotations are obtained for training and test examples, models yield approximately the same accuracy. However, in CIFAR-100, training with 20 \textit{superclasses}, the network cannot generalize well to the test examples as it achieves $97.76\%$ classification accuracy on training set but cannot go beyond $70\%$ on the test set. Therefore, accuracy of the annotations is more limited for the test examples. To monitor this effect, training set performances are also presented in Table \ref{tab:cifar}. Training/test split might be ambiguous for the semi-supervised problems discussed in this article. We would like to emphasize that coarse labels (parent-classes) are introduced to the network only for the training set examples, not for those in the test set. Fine labels, on the other hand, are never introduced to the model in any part of the training for any example.
 
\begin{table}[h!]\centering
	\ra{1.2}
	\caption{Benchmark results for the two-parent case on SVHN. The last row demonstrates the benchmark scores of the proposed framework in this article. Note that given values represent the test error.}
	
	%\resizebox{\columnwidth}{!} {
		\begin{tabular}{@{}lrr@{}}\toprule
			&SVHN $k=5$	& SVHN $k=10$ \\%	& NORB $k=5$ & NORB $k=10$ \\
			\midrule
			$k$-means	& $69.70\%\hskip 3.35em$ 	& $66.58\%\hskip 3.35em$	\\%& - 	& $50.40\%$\\
			GMM			& $70.81\%\hskip 3.35em$	& $66.73\%\hskip 3.35em$	\\%& - 	& $49.50\%$\\
			AE+$k$-means& $67.29\%\hskip 3.35em$	& $62.42\%\hskip 3.35em$	\\%& - 	& $55.40\%$\\
			AE+GMM		& $69.38\%\hskip 3.35em$	& $63.89\%\hskip 3.35em$	\\%& - 	& $51.00\%$\\
			\midrule
			ACOL		& $36.90\%(\pm 6.22)$	& $21.66\%(\pm 1.49)$	\\%& -	& $39.80\%(\pm 2.20)$ \\
			\bottomrule
			\label{tab:svhn}
		\end{tabular}
	%}
	\vskip -0.2 in
\end{table}

\begin{table}[h!]\centering
	\ra{1.2}
	\caption{Benchmark results for the twenty-parent case on CIFAR-100. The last row demonstrates the benchmark scores of the proposed framework in this article. Note that given values represent the test error.}
	
	\resizebox{\columnwidth}{!} {
	\begin{tabular}{@{}lrrrr@{}}\toprule
					&Training $k=5$	&Test $k=5$	& Training $k=10$ 	& Test $k=10$ \\
		\midrule
		$k$-means	& $65.17\%\hskip 3.35em$ 	& $64.30\%\hskip 3.35em$	& $62.04\%\hskip 3.35em$ 		& $60.55\%\hskip 3.35em$ \\
		GMM			& $65.90\%\hskip 3.35em$ 	& $64.91\%\hskip 3.35em$	& $62.72\%\hskip 3.35em$ 		& $61.33\%\hskip 3.35em$ \\
		AE+$k$-means& $64.60\%\hskip 3.35em$	& $63.81\%\hskip 3.35em$	& $61.17\%\hskip 3.35em$			& $59.63\%\hskip 3.35em$	\\
		AE+GMM		& $65.94\%\hskip 3.35em$	& $65.45\%\hskip 3.35em$	& $62.17\%\hskip 3.35em$			& $61.20\%\hskip 3.35em$\\
		\midrule
		ACOL		& $44.64\%(\pm 0.79)$ & $62.17\%(\pm 0.32)$ & $37.41\%(\pm 0.42)$ & $58.60\%(\pm 0.18)$ \\
		\bottomrule
		\label{tab:cifar}
	\end{tabular}
	}
	\vskip -0.2 in
\end{table}

\section{Conclusions}
In this paper, we introduce a novel modification to the output layer of a neural network to automatically identify the latent annotations via partial supervision of course class labels.  We use graph-based adjacency performance metrics in training the model to search for sub-classes under parent-classes without supervision.  The proposed learning framework can be used in many domains such as text categorization, protein function prediction, image classification as well as in exploratory scientific studies such as medical and genomics research. Our major contributions are four-fold:

\begin{itemize} 
	\item We explore a different type of semi-supervised setting for neural networks. That is, every observation in a dataset has a corresponding ground-truth label; however, this label is more general than the main categorization interest. Hence, the aim of this particular semi-supervised setting is to explore the more definite latent annotations when this general supervision is provided as parent-class labels.
	\item We propose a simple yet efficient output layer modification, ACOL, which enables simultaneous supervised classification and unsupervised clustering on neural networks. ACOL introduces duplicated softmax nodes for each one of the parent-classes.  
	\item We adopted GAR terms for the unsupervised portion of the objective function and showed that these terms efficiently guide the optimization in a way that each softmax duplicate is specialized during the training to represent a proper latent annotation.
	\item Most interestingly, we demonstrate that the neural network can learn from existing differences between different parent-class labels and translate that knowledge to better identify sub-classes within each parent-class.
\end{itemize}

Finally the proposed approach is validated on three popular image benchmark datasets, MNIST, SVHN and CIFAR-100, through t-SNE visualizations and unsupervised clustering accuracy metrics compared to well-accepted approaches implemented for the particular semi-supervised setting discussed in this paper.

\bibliography{bibliography}
\bibliographystyle{thisIEEEtran}

\end{document}